# Pattern Classification using Simplified Neural Networks with Pruning Algorithm

S. M. Kamruzzaman[1]
Ahmed Ryadh Hasan[2]

**Abstract:** In recent years, many neural network models have been proposed for pattern classification, function approximation and regression problems. This paper presents an approach for classifying patterns from simplified NNs. Although the predictive accuracy of ANNs is often higher than that of other methods or human experts, it is often said that ANNs are practically "black boxes", due to the complexity of the networks. In this paper, we have an attempted to open up these black boxes by reducing the complexity of the network. The factor makes this possible is the pruning algorithm. By eliminating redundant weights, redundant input and hidden units are identified and removed from the network. Using the pruning algorithm, we have been able to prune networks such that only a few input units, hidden units and connections left yield a simplified network. Experimental results on several benchmarks problems in neural networks show the effectiveness of the proposed approach with good generalization ability.

## 1 Introduction

In recent years, many neural network models have been proposed for pattern classification, function approximation and regression problems [2] [3] [18]. Among them, the class of multi-layer feed forward networks is most popular. Methods using standard back propagation perform gradient descent only in the weight space of a network with fixed topology [13]. In general, this approach is useful only when the network architecture is chosen correctly [9]. Too small a network cannot learn the problem well or too large a size will lead to over fitting and poor generalization [1].

Artificial neural networks are considered as efficient computing models and as the universal approximators [4]. The predictive accuracy of neural network is higher than that of other methods or human experts, it is generally difficult to understand how the network arrives at a particular decision due to the complexity of a particular architecture [6] [15]. One of the major criticism is their being black boxes, since no satisfactory explanation of their behavior has been offered. This is because of the complexity of the interconnections between layers and the network size [18]. As such, an optimal network size with minimal number of interconnection will give insight into how neural network performs. Another motivation for network simplification and pruning is related to time complexity of learning time [7] [8].

## 2 Pruning Algorithm

Network pruning offers another approach for dynamically determining an appropriate network topology. Pruning techniques [11] begin by training a larger than necessary network and then eliminate weights and neurons that are deemed redundant. Typically, methods for removing weights involve adding a penalty term to the error function [5]. It is hoped that adding a penalty term to the error function, unnecessary connection will have smaller weights and therefore complexity of the network can be significantly reduced. This paper aims at pruning the network size both in number of neurons and number of interconnections between the neurons. The pruning strategies along with the penalty function are described in the subsequent sections.

[1] Assistant Professor, Department of Computer Science and Engineering, Manarat International University, Dhaka-1212, Bangladesh, Email: smk_iiuc@yahoo.com

[2] School of Communication, Independent University Bangladesh, Chittagong, Bangladesh.
Email: ryadh78@yahoo.com

## 2.1 Penalty Function

When a network is to be pruned, it is a common practice to add a penalty term to the error function during training [16]. Usually, the penalty term, as suggested in different literature, is

$$P(w,v) = \varepsilon_1 \left( \sum_{m=1}^{h} \sum_{l=1}^{n} \frac{\beta(w_l^m)^2}{1+\beta(w_l^m)^2} + \sum_{m=1}^{h} \sum_{p=1}^{o} \frac{\beta(v_p^m)^2}{1+\beta(v_p^m)^2} \right) + \varepsilon_2 \left( \sum_{m=1}^{h} \sum_{l=1}^{n} (w_l^m)^2 + \sum_{m=1}^{h} \sum_{p=1}^{o} (v_p^m)^2 \right) \quad (1)$$

Given an n-dimensional example $x^i, i\varepsilon\{1,2,....,k\}$ as input, let $w_l^m$ be the weight for the connection from input unit $l, l\varepsilon\{1,2,....,n\}$ to hidden unit $m, m\varepsilon\{1,2,....,h\}$ and $v_p^m$ be the weight for the connection from hidden unit m to output unit $p, p\varepsilon\{1,2,....,o\}$, the $p^{th}$ output of the network for example $x^i$ is obtained by computing

$$S_p^i = \sigma\left(\sum_{m=1}^{h} \alpha^m v_p^m\right), \text{ where} \quad (2)$$

$$\alpha^m = \delta\left(\sum_{l=1}^{n} x_l^i w_l^m\right), \ \delta(x) = (e^x - e^{-x})/(e^x - e^{-x}) \quad (3)$$

The target output from an example $x^i$ that belongs to class $C_j$ is an o-dimensional vector $t^i$, where $t_p^i = 0$ if $p = j$ and $t_p^i = 1, j, p = 1,2…o$. The back propagation algorithm is applied to update the weights $(w, v)$ and minimize the following function:

$$\theta(w,v) = F(w,v) + P(w,v) \quad (4)$$

where $F(w,v)$ is the cross entropy function as defined

$$F(w,v) = -\sum_{i=1}^{k} \sum_{p=1}^{o} \left(t_p^i \log S_p^i + (1-t_p^i)\log(1-S_p^i)\right) \quad (5)$$

and $P(w,v)$ is a penalty term as described in (1) used for weight decay.

## 2.2 Redundant weight Pruning

Penalty function is used for weight decay. As such we can eliminate redundant weights with the following Weight Elimination Algorithm as suggested in different literature [12][14][17].

### 2.2.1 Weight Elimination Algorithm:
1. Let η$_1$ and η$_2$ be positive scalars such that η$_{1+}$ η$_2$ < 0.5.
2. Pick a fully connected network and train this network such that error condition is satisfied by all input patterns. Let (w, v) be the weights of this network.
3. For each $w_l^m$, if

$$\left|v^m \times w_l^m\right| \leq 4 \eta_2 \quad (6)$$

Then remove $w_l^m$ from the network

4. For each $v^m$, If
$$|v^m| \leq 4\eta_2 \quad (7)$$
Then remove $v^m$ from the network

5. If no weight satisfies condition (6) or condition (7) then remove $w_l^m$ with the smallest product $|v^m \times w_l^m|$.

6. Retrain the network. If classification rate of the network falls below an acceptable level, then stop. Otherwise go to Step 3.

## 2.3 Input and Hidden Node Pruning

A node-pruning algorithm is presented below to remove redundant nodes in the input and hidden layer.

### 2.3.1 Input and Hidden node Pruning Algorithm:

**Step 1:** Create an initial network with as many input neurons as required by the specific problem description and with one hidden unit. Randomly initialize the connection weights of the network within a certain range.

**Step 2:** Partially train the network on the training set for a certain number of training epochs using a training algorithm. The number of training epochs, τ, is specified by the user.

**Step 3:** Eliminate the redundant weights by using weight elimination algorithm as described in section 2.2.

**Step 4:** Test this network. If the accuracy of this network falls below an acceptable range then add one more hidden unit and go to step 2.

**Step 5:** If there is any input node $x_l$ with $w_l^m = 0$, for m = 1,2…h, then remove this node.

**Step 6:** Test the generalization ability of the network with test set. If the network successfully converges then erminate, otherwise, go to step 1.

## 3. Experimental Results And Discussions

In this experiment, we have used three benchmark classification problems. The problems are breast cancer diagnosis, classification of glass types and Pima Indians Diabetes diagnosis problem [10] [19]. All the data sets were obtained from the UCI machine learning benchmark repository. Brief characteristics of the data sets are listed in Table 1.

**Table 1:** Characteristics of data sets.

| Data set | Input Attributes | Output Units | Output Classes | Training Examples | Validation Examples | Test examples | Total examples |
|---|---|---|---|---|---|---|---|
| **Cancer1** | 9 | 2 | 2 | 350 | 175 | 174 | 699 |
| **Glass** | 9 | 6 | 6 | 107 | 54 | 54 | 215 |
| **Diabetes** | 8 | 2 | 2 | 384 | 192 | 192 | 768 |

The experimental results of different data sets are shown in table 2, figure 1, 2 and 3. In the experimental results of cancer data set, we have found that a fully connected network of 9-3-2 architecture has the classification accuracy of 97.143%. After pruning the network with *Weight Elimination Algorithm* and *Input and Hidden node Pruning Algorithm,* we have found a simplified network of 3-1-2 architecture with classification accuracy of 96.644%.
The graphical representation of the simplified network is given in figure 3. It shows that only the input attributes $I_1$, $I_6$, $I_9$ along with a single hidden unit is adequate for this problem.

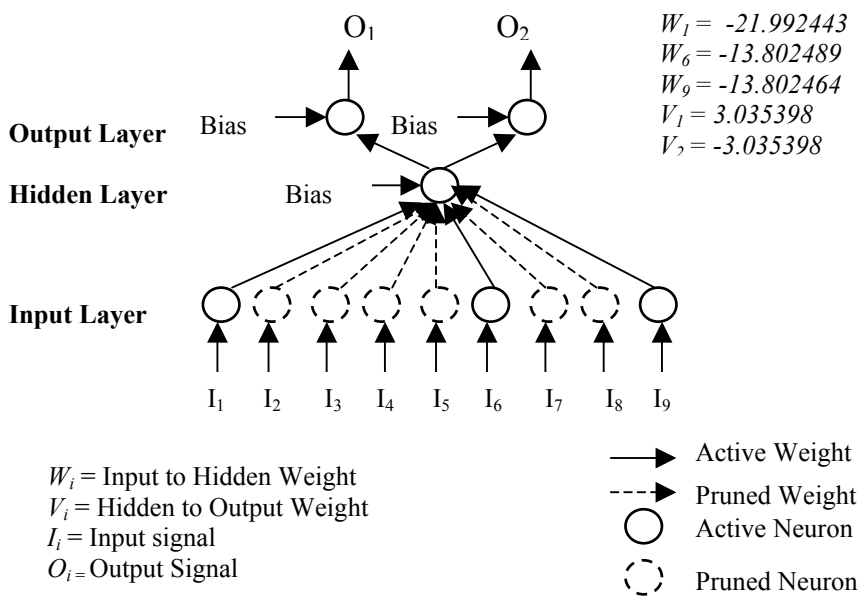

**Figure 1:** Simplified Network for Breast Cancer Diagnosis problem.

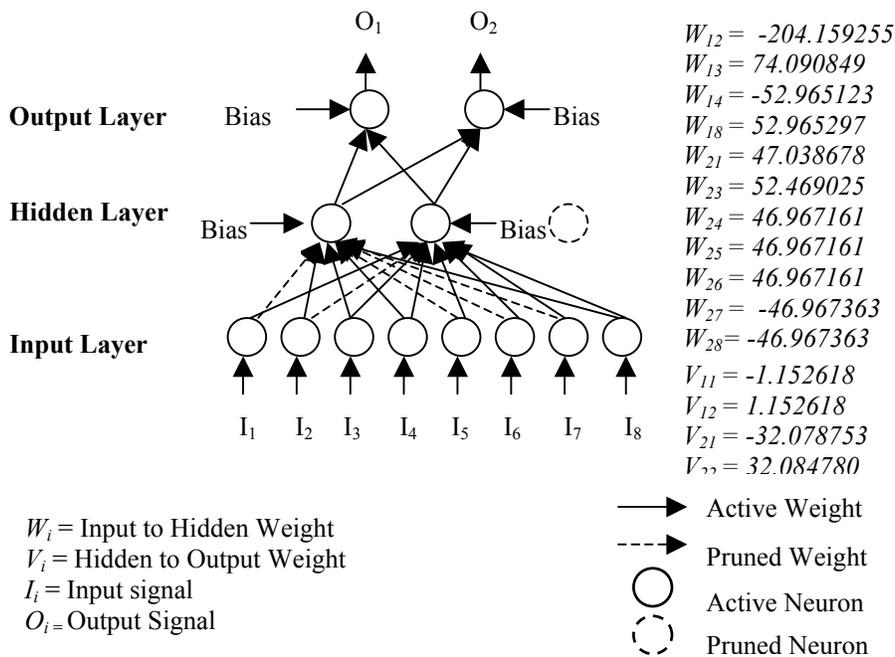

**Figure 2:** Simplified Network for Pima Indians Diabetes diagnosis problem.

In the experimental results of Pima Indians Diabetes data set, we have found that a fully connected network of 8-3-2 architecture has the classification accuracy of 77.344%. After pruning the network with *Weight Elimination Algorithm* and *Input and Hidden node Pruning Algorithm,* we have found a simplified network of 8-2-2 architecture with classification accuracy of 75.260%. The graphical representation of the simplified network is given in figure 2. It shows that no input attribute can be removed but a hidden node along with some redundant connection has been removed which have been shown with a dotted line in figure 2.

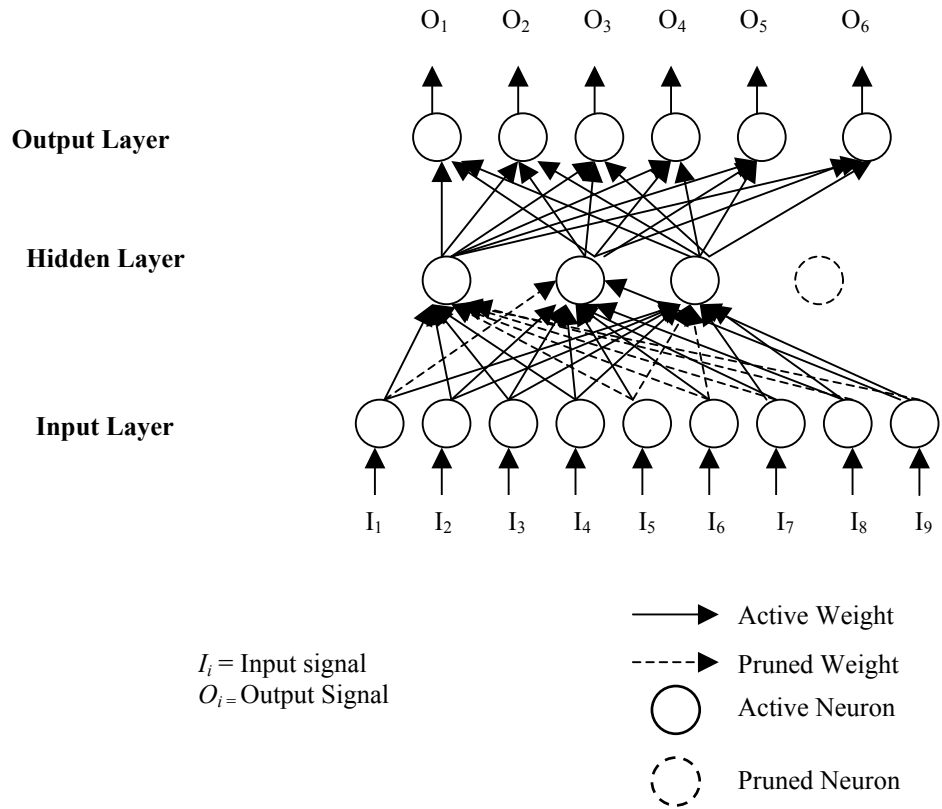

**Figure 3:** Simplified Network for Glass classification problem.

In the experimental results of Glass classification data set, we have found that a fully connected network of 9-4-6 architecture has the classification accuracy of 65.277%. After pruning the network with *Weight Elimination Algorithm* and *Input and Hidden node Pruning Algorithm,* we have found a simplified network of 9-3-6 architecture with classification accuracy of 63.289%. The graphical representation of the simplified network is given in figure 3. It shows that no input attribute can be removed but a hidden node along with some redundant connection has been removed which have been shown with a dotted line in figure 3.

**Table 2:** Experimental Results

| Data sets<br>Results | Cancer1 | Diabetes | Glass |
|---|---|---|---|
| **Learning Rate** | 0.1 | 0.1 | 0.1 |
| **No. of Epoch** | 500 | 1200 | 650 |
| **Initial Architecture** | 9-3-2 | 8-3-2 | 9-4-6 |
| **Input Nodes Removed** | 6 | 0 | 1 |
| **Hidden Nodes Removed** | 2 | 1 | 2 |
| **Total Connection Removed** | 24 | 13 | 16 |
| **Simplified Architecture** | 3-1-2 | 8-2-2 | 9-3-6 |
| **Accuracy (%) of fully connected network** | 97.143 | 77.344 | 65.277 |
| **Accuracy (%) of simplified network** | 96.644 | 75.260 | 63.289 |

From the experimental results discussed above, it can be said that not all input attributes and weights are equally important. Moreover, it is difficult to determine the appropriate number of hidden nodes. By pruning approach we can automatically determine an appropriate number of hidden nodes. We can remove redundant nodes and connections without sacrificing significant accuracy using network pruning approach discussed in section 2.2 and section 2.3. As such we can reduce computational cost by using the simplified networks.

## 4 Future Work

In future we will use this network pruning approach for rule extraction and feature selection. These pruning strategies will be also examined for function approximation and regression problems.

## 5 Conclusions

In this paper we proposed an efficient network simplification algorithm using pruning strategies. Using this approach we obtain optimal network architecture with minimal number of connections and neurons without deteriorating the performance of the network significantly. Experimental results show that the performance of the simplified network is quite significant and acceptable compared to fully connected network. This simplification of the network ensures both reliability and reduced computational cost.